\documentclass{Interspeech}

\usepackage{tipa}
\usepackage{xcolor}
\usepackage{colortbl}
\usepackage{multirow}
\usepackage[toc,acronyms]{glossaries}
\usepackage{comment}
\usepackage{hyperref}

\interspeechcameraready

\title{\dos: Dementia Obfuscation in Speech}

\newcommand{\dos}{ClaritySpeech}

\author[affiliation={1}]{Dominika}{Woszczyk}
\author[affiliation={1,2}]{Ranya}{Aloufi}
\author[affiliation={1}]{Soteris}{Demetriou}

\affiliation{}{Imperial College London}{UK}
\affiliation{}{Taibah University} {Saudi Arabia}
\email{d.woszczyk19@imperial.ac.uk, r.aloufi18@imperial.ac.uk, s.demetriou@imperial.ac.uk}

\definecolor{lightgreen}{HTML}{e5f4d5}
\definecolor{lightblue}{HTML}{a2d2ff}

\begin{document}

\maketitle

\begin{abstract}
    

Dementia, a neurodegenerative disease, alters speech patterns, creating communication barriers and raising privacy concerns. Current speech technologies, such as automatic speech transcription (ASR), struggle with dementia and atypical speech, further challenging accessibility. This paper presents a novel dementia obfuscation in speech framework, ClaritySpeech, integrating ASR, text obfuscation, and zero-shot text-to-speech (TTS) to correct dementia-affected speech while preserving speaker identity in low-data environments without fine-tuning. Results show a 16\% and 10\% drop in mean F1 score across various adversarial settings and modalities (audio, text, fusion) for ADReSS and ADReSSo, respectively, maintaining 50\% speaker similarity. We also find that our system improves WER (from 0.73 to 0.08 for ADReSS and 0.15 for ADReSSo) and speech quality from 1.65 to $\sim$ 2.15, enhancing privacy and accessibility.\footnote{https://github.com/domiwk/clarityspeech}\footnote{Samples are available at https://stereomousse.github.io.}
 
\end{abstract}

\section{Introduction}

As dementia develops, it alters speech patterns, introducing disfluencies, pauses, lexical swaps, and convoluted grammar~\cite{gosztolya2019identifying, konig2015automatic, ahmed2013connected}. These changes not only create communication challenges but also become identifiable markers of the individual's health condition, raising privacy risks and limiting access to speech-operated services~\cite{halpern2024quantifying, de2020artificial}. Individuals with dementia are exposed through their voices in both professional and daily life, including identity verification (e.g., voice-authenticated banking) and platforms like podcasts, YouTube, and social media. As speech interfaces like voice assistants and self-driving cars become more common, preserving privacy while maintaining clear communication for dementia patients is increasingly important. Exposed individuals may be flagged as vulnerable, leading to potential discrimination by employers or financial services, or exploitation by malicious agents.

Additionally, despite advances in \newacronym{asr}{ASR}{automatic speech transcription}\gls{asr}~\cite{baevski2020wav2vec,radford2023robust} current systems face significant limitations in processing dementia-affected and other unusual speech~\cite{qian2023survey}. These systems often struggle to accurately transcribe disordered speech because they are predominantly trained on fluent and clear speech data, resulting in high transcription errors~\cite{ qian2023survey, likhomanenko2020rethinking}.
A naive solution is to resynthesize one's voice to improve its quality but this does not take into consideration the linguistic properties of dementia speech. 

Furthermore, existing \newacronym{tts}{TTS}{text-to-speech}\gls{tts} models~\cite{casanova2024xtts,peng2024voicecraft}, trained on clean and fluent data, need to be adapted to atypical patterns which require a significant amount of audio recordings to train on. This is both difficult for dementia patients and creates a possible privacy leakage if the models are trained in the cloud. Finally, restoring and preserving one's original speaker timbre has a crucial impact on an individual's psyche and identity~\cite{mertl2018quality}.



\textbf{Prior Works} Privacy-preserving techniques like anonymization and attribute obfuscation focus on acoustic features and recently approaches such as adversarial techniques and style transfer have been applied, but are challenging with low-resource datasets~\cite{noe2020adversarial, chouchane2023differentially, aloufi2020privacy}, which is often the case with disordered speech. These systems are also not concerned with transforming the content to preserve privacy, while dementia affects both acoustic and linguistic characteristics of speech. Speech editing systems, such as FluentSpeech~\cite{jiang2023fluentspeech} or Voicecraft~\cite{peng2024voicecraft}, can modify speech locally to correct disfluencies, accents or stutters. Dysarthric speech shares some acoustic features and low data availability with dementia speech.  However, an important distinction is that dementia is a cognitive disorder primarily affecting memory, thinking, and social abilities, while dysarthria specifically affects the physical production of speech. Techniques such as data augmentation ~\cite{soleymanpour2024accurate,vachhani18_interspeech,hermann2023few}, HMM-based synthesis~\cite{veaux2012using, creer2013building} and voice conversion (VC)~\cite{huang2021preliminary, wang2020learning} have been applied but focus on acoustic features and do not target content.

\textbf{Our Approach} In this paper, we propose a framework that utilizes the normalizing nature of zero-shot \gls{tts} synthesis and its training-free setting with a text-based obfuscation mechanism aimed specifically at dementia-related patterns.  Unlike existing systems that focus on anonymization, noise reduction or disfluency correction, our method obfuscates cognitive impairment patterns while also preserving speaker identity. We evaluate multiple zero-shot \gls{tts} models, assessing their ability to minimize dementia leakage (privacy) and the impact of our framework on the \gls{asr} utility task, speaker similarity, and speech quality. We outline our contributions below:


\begin{itemize}
\item \textbf{Holistic Dementia Obfuscation Framework} We propose a novel speech obfuscation framework that tackles cognitive markers in dementia speech while improving its fluency and quality and preserving the speaker's identity.

\item \textbf{Low-resource Setting} Our system works in a zero-shot setting and does not require fine-tuning.

\item \textbf{Rigorous Empirical Evaluation} We investigate the privacy (dementia leakage) and utility (ASR, speaker similarity, audio quality) of the proposed framework. 

\item \textbf{Improved Accessibility} We find that our system improves fluency and utility for tasks such as \gls{asr}.

\end{itemize}

\section{\dos}
\label{sec:dos_meth}

\subsection{Threat and Defense Models}

We consider an adversary with access to a target user's recorded speech with the goal of detecting whether the target user suffers from dementia. The adversary can operate in two settings: (a) a static setting and (b) an adaptive setting. A static adversary ($\mathcal{A}$) is obfuscation-oblivious and has direct access to the raw audio. An adaptive adversary ($\mathcal{A}_{Ada}$) is obfuscation-aware. We assume that $\mathcal{A}_{Ada}$ has gained access to samples generated by any applied obfuscation strategy and can leverage these obfuscated samples to adapt and improve their detection capabilities.



\subsection{Dementia Obfuscation in Speech}

\begin{figure}[!ht]
    \centering
    \includegraphics[width=\linewidth]{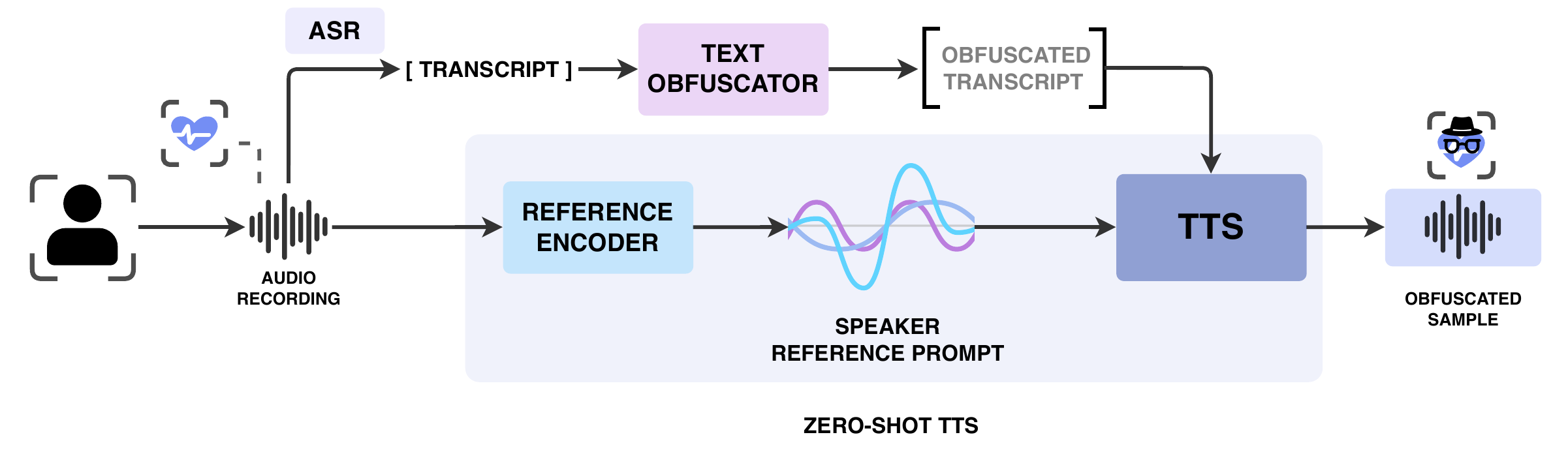}
    \caption{End-to-end ClaritySpeech framework. The input speech is transcribed through \gls{asr}, obfuscated via a text obfuscator and fed to the zero-shot \gls{tts} system together with the reference speech, which outputs the obfuscated sample in the reference voice.}
    \label{fig:dos_framework}
\end{figure}

We present an end-to-end framework for dementia obfuscation in speech, \dos, consisting of three components: an automatic speech recognition (ASR) module, a text obfuscation mechanism, and a zero-shot text-to-speech (TTS) system. The framework conceals dementia-related attributes while preserving the speaker's voice and improving intelligibility and naturalness. We describe each component below.



\vspace{3pt}\noindent\textbf{Automatic Speech Recognition} The first component of our framework is the automatic speech recognition step, which transcribes the audio recording to obfuscate its content. The main challenge of this module is transcription accuracy. Speech affected by dementia can exhibit characteristics such as hesitations, mispronunciations, and non-standard, sometimes incoherent, phrasing making it difficult for ASR systems to make correct predictions. We choose Whisper large v3~\cite{radford2023robust} as our \gls{asr} model, currently the \newacronym{sota}{SOTA}{state-of-the-art}\gls{sota}. Whisper has demonstrated robust capabilities in transcribing speech across diverse acoustic environments and speaker conditions. We pick a robust \gls{asr} system to minimize the transcription rate even in our atypical setting. The impact of this step on privacy and utility will be analyzed in our experiments section.

\vspace{3pt}\noindent\textbf{Dementia Obfuscation in Text} Once the content of the speech is transcribed into a text form, the next step involves applying the text obfuscation mechanism. This is the core privacy-preserving step of the framework. The challenges here consist of preserving the original meaning and correcting for incoherence, on top of concealing dementia-specific characteristics and maintaining naturalness. We base our system on DiDOTS ~\cite{woszczyk2025didots}, which was designed to modify the content of the transcribed text to reduce or eliminate linguistic markers linked with dementia. Through knowledge distillation with a large language model, the model learned to effectively paraphrase sentences and obfuscate dementia while preserving the meaning and intent of the original text. The model has been extensively studied for its impact on dementia leakage and semantic preservation. 

\vspace{3pt}\noindent\textbf{Zero-Shot Text-to-Speech} The final step in the framework is to synthesize obfuscated speech using a \gls{tts} system. We propose to use a zero-shot TTS model that was trained on publicly available data. Current Zero-shot TTS achieve great voice cloning capabilities and can adapt to new voices with minimal data and short reference segments (6s). The reference speech is encoded and used to condition the generation at inference time. This allows us to both not train the model given the low amount of data per speaker but also not to introduce further privacy vulnerabilities. Furthermore, we hypothesize that these models have a normalizing nature. Indeed, zero-shot models work best when trained on very diverse data so they can learn to re-create features seen during training. However, they struggle to reproduce atypical speech (young, old, accents, dysarthric, ...). We want to utilize this ``weakness" to normalize dementia speech. We believe that while these models can transfer a general timbre, they will often smooth out granularities such as tempo or unstable pitch and volume.
In our experimentation, we compare SOTA models and their impact on the detection abilities of dementia classifiers and acoustic features.

\section{Experimental Setup}
\label{sec:dos_exp}

\begin{table*}[!ht]
\caption{Privacy/Utility evaluation of various zero-shot \gls{tts} systems on the ADReSS and ADReSSo datasets. We report the mean F1 score across static and adaptive adversaries across each modality.  The best results for each column are marked in bold. Non-significant results for the static and adaptive adversaries are marked as $\dagger$ and $*$ respectively.}
\centering
\label{tab:dos_privacy}
\resizebox{\textwidth}{!}{%
\begin{tabular}{ll|lll|lll|lll}
\toprule
\textbf{Dataset} &\textbf{System} & \textbf{Audio↓} & \textbf{Text↓} & \textbf{Fusion↓} & \textbf{Mean Static↓} & \textbf{Mean Adaptive↓} & \textbf{Total Mean↓} & \textbf{SS↑} & \textbf{WER↓} & \textbf{UTMOS↑} \\
\midrule
 \multirow{3}{*}{ADReSS}  & Original & 0.64& 0.72& 0.73& \multicolumn{2}{c}{0.7} & \cellcolor[HTML]{DAE8FC}0.7  & - & 0.73&1.65\\
 \cmidrule{2-11} 
  & ClaritySpeech w/ XTTSv2 & 0.55$\dagger$& \textbf{0.59}& \textbf{0.58}& \textbf{0.49} & \textbf{0.65} & \cellcolor[HTML]{DAE8FC}\textbf{0.59}  & 0.50 & 0.08 &2.15 \\
 & \hspace{1.8cm} w/ StyleTTS2 & \textbf{0.54}$*$ & 0.61
& 0.62
& 0.51 & 0.66 & \cellcolor[HTML]{DAE8FC}0.61  & 0.36 & \textbf{0.04} &\textbf{2.86} \\
 & \hspace{1.8cm} w/ Hierspeech++ & 0.56$*$ & 0.64$*$& 0.61
& 0.52 & 0.68 & \cellcolor[HTML]{DAE8FC}0.63  & \textbf{0.52} & 0.21 &2.46 \\
\midrule
  \multirow{3}{*}{ADReSSo} & Original & 0.55
& 0.67& 0.66& \multicolumn{2}{c}{0.63} & \cellcolor[HTML]{DAE8FC}0.63  & - & - &1.60\\
 \cmidrule{2-11} 
  & ClaritySpeech w/ XTTSv2 & \textbf{0.51}& 0.58$*$& 0.54& 0.51 & 0.58 & \cellcolor[HTML]{DAE8FC}0.56 & \textbf{0.53} & 0.15 &2.13 \\
  & \hspace{1.8cm} w/ StyleTTS2 & 0.54$*$& 0.58& \textbf{0.53}& 0.54 & \textbf{0.55} & \cellcolor[HTML]{DAE8FC}\textbf{0.55}  & 0.39 & \textbf{0.08} &\textbf{2.68} \\
 & \hspace{1.8cm} w/ Hierspeech++ & 0.55$\dagger$& \textbf{0.56}& \textbf{0.53}& \textbf{0.50} & 0.58 & \cellcolor[HTML]{DAE8FC}\textbf{0.55}  & 0.51 & 0.14 &2.62 \\
 \bottomrule
\end{tabular}%
}
\end{table*}

\begin{table*}[!ht]
\caption{Impact of the different \dos~components on the privacy and utility metrics of various zero-shot \gls{tts} systems, on the ADReSS dataset. We report the mean F1 score across static and adaptive adversaries across each modality. The best results for each column are marked in bold. Non-significant results for the static, adaptive adversaries and utility are marked as $\dagger$, $*$ and $\nabla$ respectively.}
\centering
\small
\label{tab:dos_ablation}
\resizebox{\textwidth}{!}{%
\begin{tabular}{@{}llll|lll|lll@{}}
\toprule
\textbf{System} & \textbf{Audio↓} & \textbf{Text↓} & \textbf{Fusion↓} & \textbf{Mean Static↓} & \textbf{Mean Adaptive↓} & \textbf{Total Mean↓} & \textbf{SS↑} & \textbf{WER↓} & \textbf{UTMOS↑} \\
\midrule
\dos & 0.55 & 0.59 & 0.57 & 0.49 & 0.65 & \cellcolor[HTML]{DAE8FC}0.57& 0.50 & 0.08 & 2.15 \\
\midrule
w/o ASR & \textbf{0.47}$\dagger$& \textbf{0.50}$*$& 0.63 & \textbf{0.44}& \textbf{0.63}& \cellcolor[HTML]{DAE8FC}\textbf{0.54}& 0.48$\nabla$& 0.08$\nabla$& 2.21 \\
w/o Text obfuscation & 0.53$\dagger$ & 0.68$*$ & 0.66 & 0.56 & 0.67 & \cellcolor[HTML]{DAE8FC}0.62& \textbf{0.56}& 0.23 & 2.15$\nabla$\\
w/o ZS & 0.57$*$ & 0.60$*$ & \textbf{0.56}$\dagger$$*$& 0.48 & 0.65 & \cellcolor[HTML]{DAE8FC}0.57& 0.12 & \textbf{0.02}& \textbf{2.93}\\
w/o ZS \& Text obfuscation & 0.53$\dagger$ & 0.69$\dagger$$*$ & 0.68 & 0.60 & 0.68 & \cellcolor[HTML]{DAE8FC}0.64& 0.16 & 0.29 & 2.91\\
\bottomrule
\end{tabular}%
}
\end{table*}

\subsection{Datasets}
\label{sec:dos_datasets}

We evaluate our systems on the \newacronym{adr}{ADR}{ADReSS}\gls{adr}~\cite{luz2020alzheimer} and \newacronym{adro}{ADRo}{ADReSSo}\gls{adro}~\cite{luz2020alzheimer} datasets. We split our datasets into sentence-level segments. For ADReSS, manual transcripts are provided and we split each sample into segments using the sentence-level timestamps from the transcripts. We split the segments with an 80-20\% split for train and test. We further process samples with voice activity detection~\cite{Silero_VAD} to trim silences at the beginning and end of segments and remove samples shorter than 3s. We end up with 448 samples (243 CC $|$ 205 AD) in the training set and 242 (140 CC $|$ 102 AD) in the test and validation sets. For ADReSSo, we use the provided segmentation timestamps to isolate segments spoken by the patients, filter them, and get 705 samples in the train set (360 CC $|$ 345 AD) and 242 samples (191 CC $|$ 146 AD) in the test and validation sets.

\subsection{Adversarial Models}

\vspace{3pt}\noindent\textbf{Models} We select a detection model for each modality, audio and text, as well as their fusion, for our dementia classifiers (adversaries). For the text-based detection model, we re-implement a BERT model based on ~\cite{yuan2020disfluencies} and \cite{hledikova2022data} with a learning rate of 1e-6, 10
epochs with early stopping (patience =1 on validation loss), gradient
clipping of 1, input length limit of 256 tokens, and batch size of
8, trained on augmented data through back translation. For the acoustic-based detection model, we base it on a pre-trained wav2vec model~\cite{baevski2020wav2vec}, which is a self-supervised model which has shown \gls{sota} performance on the DementiaBank datasets~\cite{agbavor2022artificial}. We extract embedding with the wav2vec audio model and feed it to a linear classifier (three layers and a dropout of 0.1) for binary classification. For the fusion of both, we perform early fusion and concatenate BERT and wav2vec embeddings before feeding it to the linear classifier.

\vspace{3pt}\noindent\textbf{Static and Adaptive Settings}  We evaluate our framework in two settings: static and adaptive. Under the static scenario, the adversary has only access to raw data while in the static scenario, the models have knowledge of the obfuscation mechanism used and are trained on both raw and obfuscated samples.

\subsection{TTS Systems}


\vspace{3pt}\noindent\textbf{XTTSv2} was introduced for the task of zero-shot multilingual speech synthesis and achieves \gls{sota} performance in terms of voice cloning~\cite{casanova2024xtts}. XTTSv2 trains a GPT-based language model for text and audio token predictions conditioned on speaker prompt tokens. The audio codecs are decoded via a HiFiGAN~\cite{kong2020hifi} and trained with speaker similarity loss. The
model is accessed through the coqui-tts package~\footnote{https://github.com/idiap/coqui-ai-TTS}. We pick XTTSv2 as our chosen \gls{tts}  model given its great cloning abilities but also unlike diffusion and flow-based models, it does not rely on fine-grain prosody control but a reference encoder that learns to transfer the acoustic and stylistic features automatically.

\vspace{3pt}\noindent\textbf{StyleTTS 2} ~\cite{li2024styletts} is a diffusion-based zero-shot voice cloning system trained through adversarial learning which has also achieved \gls{sota} results and naturalness on style transfer and voice cloning. We use the model trained on the LibriTTS train-clean-460 subset~\footnote{https://github.com/yl4579/StyleTTS2} and perform inference with the default parameters of $\alpha$ = 0.3, $\beta$ = 0.7 and diffusions steps = 5. 

\vspace{3pt}\noindent\textbf{Hierspeech++} ~\cite{lee2023hierspeech++} is a \gls{sota} model based on a hierarchical VAE, which decouples f0 and durations for more fine-grain control and better style transfer. A denoiser (MP-SENet~\cite{lu2023mp}) is also applied to the samples before extracting features for synthesis and combined with the original style features through a denoising ratio. We use the pretrained checkpoints trained on both English and Korean datasets (2 796 hours)~\footnote{https://github.com/sh\-lee\-prml/HierSpeechpp} with default parameters and we synthesize the audio samples in 24kHz.




\subsection{Evaluation Metrics}

In our evaluation, we address four research questions: (RQ1) the privacy gain in terms of dementia leakage, (RQ2) the impact of the ASR system, (RQ3) the effect of the zero-shot TTS system, and (RQ4) the efficiency of each module. To assess \textbf{privacy}, we measure the drop in adversarial F1-score for both static and adaptive settings. For \textbf{utility}, we compute the word error rate (WER) for ASR, speech quality using the UTMOS score~\cite{saeki2022utmos}, and speaker similarity (SS) through cosine similarity of ECAPA-TDNN embeddings~\cite{desplanques2020ecapa}. Finally, we evaluate \textbf{efficiency} by measuring latency in seconds and computing the real-time factor (processing time/audio length).

\section{Results}
\label{sec:ppsp_results}

\begin{table*}[!ht]
\caption{Impact of \dos~on prosody features on ADReSS$_{GT}$ (with ground truth transcription), ADReSS$_{ASR}$ and ADReSSo test sets. Statistically significant changes are highlighted.}
\label{tab:dos_prosody features}
\resizebox{\textwidth}{!}{%
\begin{tabular}{@{}lll|l|ll@{}}
\toprule
\multicolumn{1}{c}{\multirow{2}{*}{\textbf{Features}}} & \multicolumn{2}{c}{\textbf{ADReSS$_{\textbf{GT}}$}}                                                       &                                                   \textbf{ADReSS$_{\textbf{ASR}}$}& \multicolumn{2}{c}{\textbf{ADReSSo}}                                                      \\
\multicolumn{1}{c}{}                          & original                                & \dos                                   & \dos                                  & original                                & \dos                                   \\
\toprule
Syllables num.↑                                         & 
15.69 ± 8.85      & \cellcolor{lightgreen}13.17 ± 6.30     & \cellcolor{lightgreen}11.11 ± 6.67     & 16.22 ± 7.88      & \cellcolor{lightgreen}11.12 ± 5.14     \\
Energy↑                                           & $1.13 \times 10^3 \pm 1.15 \times 10^3$ & \cellcolor{lightgreen}$4.99 \times 10^3 \pm 7.73 \times 10^2$ & \cellcolor{lightgreen}$4.96 \times 10^3 \pm 9.24 \times 10^2$ & $2.43 \times 10^3 \pm 1.86 \times 10^3$ & \cellcolor{lightgreen}$4.78 \times 10^3 \pm 7.82 \times 10^2$ \\
Pause length↓                                      & 438.73 ± 399.55   & \cellcolor{lightgreen}373.35 ± 174.59  & 401.98 ± 198.10                        & 386.47 ± 206.60   & \cellcolor{lightgreen}463.87 ± 256.78  \\
F0$_{mean}$↑                                        & 92.24 ± 44.11     & \cellcolor{lightgreen}106.40 ± 34.45   & \cellcolor{lightgreen}100.37 ± 35.18   & 102.37 ± 40.27    & \cellcolor{lightgreen}94.73 ± 31.90    \\
Pause num.↓                                        & 1.66 ± 1.36       & \cellcolor{lightgreen}0.69 ± 0.91      & \cellcolor{lightgreen}0.60 ± 0.84      & 1.36 ± 1.28       & \cellcolor{lightgreen}0.86 ± 0.99      \\
Jitter↓                                        & 0.03 ± 0.01       & \cellcolor{lightgreen}0.03 ± 0.01      & \cellcolor{lightgreen}0.03 ± 0.01      & 0.02 ± 0.01       & \cellcolor{lightgreen}0.03 ± 0.01      \\
Shimmer↑                                        & 0.13 ± 0.05                             & 0.12 ± 0.04                            & 0.13 ± 0.04                            & 0.12 ± 0.04                             & 0.12 ± 0.03                            \\
Speech rate                       & 2.77 ± 1.08       & \cellcolor{lightgreen}3.05 ± 0.76      & 2.73 ± 0.92                            & 3.02 ± 0.92       & \cellcolor{lightgreen}2.64 ± 0.80 \\
\bottomrule
\end{tabular}%
}
\end{table*}

\vspace{5pt}\noindent\textbf{Privacy \& Utility Evaluation}
We evaluate our proposed system for privacy and utility with different zero-shot \gls{tts} systems and report the results for the ADR and ADRo datasets in Table~\ref{tab:dos_privacy}. We find that all systems provide significant privacy and utility gain regardless of the TTS system\footnote{We perform a McNemar test for the adversaries and Mann-Whitney U for utility scores.}. However, they exhibit tradeoffs which showcase the importance of selecting the right TTS. On the ADR dataset, XTTSv2 achieves the largest decrease in dementia detection (0.59), outperforming StyleTTS2 (0.61) and Hierspeech++ (0.63). It consistently achieves the best results across all modalities and adversaries (audio static: 0.47; text adaptive: 0.66).
StyleTTS2 and Hierspeech++ perform comparably, with StyleTTS2 having slightly better results for static adversaries in audio (0.46 vs. 0.45) but higher leakage overall. On ADRo, StyleTTS2 and Hierspeech++ share the best total mean score (0.55), with XTTSv2 slightly behind (0.56). Hierspeech++ performs the best in text and fusion static scores (0.55 and 0.45, respectively), while StyleTTS2 performs better in adaptive fusion (0.57). We note that adaptive adversaries are quite strong, and we hypothesize that given the noisy nature of the original samples, the addition of the cleaner synthesized samples might facilitate learning detailed features. Regarding utility and audio quality, StyleTTS2 achieves the highest utility scores, but it does poorly at speaker similarity (0.36 and 0.39 on ADR, ADRo). Indeed, if the model does not transfer dementia characteristics, its utility and audio quality will improve. Hierspeech++ balances privacy and utility, offering high speaker similarity and moderate privacy leakage across datasets, but also the highest WER (0.21).
We find that XTTSv2 strikes the best balance for privacy and utility and speaker characteristics transfer, affirming our design choice.

\vspace{5pt}\noindent\textbf{Ablation Evaluation} To better understand the impact of each element of our design and ASR, we perform an ablation study and evaluate the sub-systems against privacy and utility metrics. The ASR system seems to have a slight negative impact on the privacy metrics with a mean difference of 3\% across all adversaries. However, it has little to no impact on the utility.  The text obfuscation module effectively reduces the privacy leakage in Text by 10\% and the overall mean by 5\%, but reduces speaker similarity, possibly due to the content mismatch. Generating the raw transcribed text from ASR also introduces disfluencies in speech harder to transcribe later on, as indicated by the higher WER. Unsurprisingly, the w/o ZS setting achieves the highest UTMOS value and lowest WER, but also the lowest speaker similarity as the target voice is unrelated. However, the w/o ZS \& Text obfuscation while reaching high audio quality values suffers from poor WER, indicating the importance of the text sanitization step beforehand. We note that the original WER of the ADReSS dataset after transcription is 0.73 (See Table~\ref{tab:dos_privacy}), which all of these systems improve on significantly.

\vspace{5pt}\noindent\textbf{Feature Analysis} Next, we investigate the effect of our framework on the acoustic features of obfuscated samples. We extract features as noted by the literature to be key for identifying dementia in speech and report the mean for the original and obfuscated samples in Table~\ref{tab:dos_prosody features}. We find that \dos~ significantly modifies prosody features, reducing pauses and increasing pitch (F0), energy, and speech rate. There are however minimal changes in jitter and shimmer. Surprisingly, \dos~ decreases the number of syllables, which could be attributed to the shortening tendencies of the text obfuscation mechanism, making sentences more coherent and concise. The difference between ADR$_{GT}$ and ADR$_{ASR}$ does indicate the importance of the content and its impact on the acoustics. 



\vspace{5pt}\noindent\textbf{Efficiency Evaluation} We report the latency of each of the components of our proposed framework in Table~\ref{tab:dos_rtf} on a CPU. We find that the overall system has a non-negligible processing time (average of 11s for samples in the range 3-6s) with a mean real-time factor of 6.07. However, we find that most of this overhead comes from the TTS and ASR  modules which on their own introduce a lag of circa 3s and 2s per second of speech respectively. On the other hand, text obfuscation is a relatively lightweight component which is 4x times faster than real-time. 

\begin{table}[]
\caption{Mean and standard deviation of inference time and real-time-factor (RTF) for each \dos~component across 100 samples on CPU.}
\centering
\small
\label{tab:dos_rtf}
\resizebox{0.75\columnwidth}{!}{%
\begin{tabular}{@{}lll@{}}
\toprule
\textbf{System} & \textbf{Mean Time in s (std)} & \textbf{Mean RTF (std)}\\
\midrule
\dos & 11.52 (3.94) & 6.07 (4.01)  \\
\midrule
ASR & 5.15 (0.74) & 3.01 (2.57)  \\
Text obfuscation & 0.56 (0.35) & 0.25 (0.16)  \\
TTS & 6.34 (3.43) & 1.71 (0.16) \\
\bottomrule
\end{tabular}%
}
\end{table}


\section{Discussion \& Limitations}

We find that the proposed framework successfully reduces privacy leaks while improving downstream utility, and zero-shot TTS can indeed correct atypicality in this setting. We also find that the impact of the transcription error is mitigated by the text obfuscation mechanism. Limitations of this work include the small, low-diversity datasets used. Future work could combine multiple dementia datasets in English or across languages to improve generalizability. Due to dataset size and quality, we did not fine-tune models in this study. Comparing zero-shot and speaker-dependent models, despite privacy concerns, would be a valuable direction for future research. Finally, future work could also focus on optimizing ASR and TTS for better real-time performance.
The proposed framework can be extended to conditions requiring both text and audio modifications (obfuscating speech in recordings for social media, storage, or live obfuscation), but the framework must be tuned for attribute-specific characteristics. 
Besides privacy, the framework could improve accessibility for individuals with stuttering or disfluent speech, enhancing speech-to-text tools, speech-based services, e-learning platforms, and lecture transcriptions.

\section{Conclusion}
This work addresses the privacy leakage of dementia in speech. We propose an end-to-end framework that transforms both text and audio to conceal dementia markers while maintaining content and speaker similarity. Our approach leverages zero-shot TTS, which filters certain acoustic features and smooths out dementia traits, enhancing downstream tasks. We demonstrate improvements in ASR utility tasks with reduced WER and improved speech quality. Our ablation study also highlights the importance of zero-shot TTS combined with obfuscated text for preserving speaker similarity and fluency. Nevertheless, the privacy and accessibility gains come with the trade-off of diminished speaker similarity. This work lays the foundation for future applications in healthcare and privacy with future work focusing on efficiency and better transfer of speaker identity.


\bibliographystyle{IEEEtran}
\bibliography{main}

\end{document}